\documentclass[10pt,twocolumn,letterpaper]{article}

\usepackage[pagenumbers]{cvpr} 

\usepackage{graphicx}
\usepackage{amsmath}
\usepackage{amssymb}
\usepackage{booktabs}
\usepackage{multirow}

\usepackage{paralist} 

\usepackage[pagebackref,breaklinks,colorlinks]{hyperref}

\usepackage[capitalize]{cleveref}
\crefname{section}{Sec.}{Secs.}
\Crefname{section}{Section}{Sections}
\Crefname{table}{Table}{Tables}
\crefname{table}{Tab.}{Tabs.}

\def\cvprPaperID{****} 
\def\confName{CVPR}
\def\confYear{2023}

\begin{document}

\title{Probabilistic Shape Completion by Estimating Canonical Factors with Hierarchical VAE}

\author{
	Wen Jiang \quad  Kostas Daniilidis \vspace{2mm} \\
	University of Pennsylvania
}

\maketitle

\begin{abstract}
We propose a novel method for 3D shape completion from a partial observation of a point cloud.
Existing methods either operate on a global latent code, which limits the expressiveness of their model, or autoregressively estimate the local features, which is highly computationally extensive.
Instead, our method estimates the entire local feature field by a single feedforward network by formulating this problem as a tensor completion problem on the feature volume of the object. Due to the redundancy of local feature volumes, this tensor completion problem can be further reduced to estimating the canonical factors of the feature volume.
A hierarchical variational autoencoder (VAE) with tiny MLPs is used to probabilistically estimate the canonical factors of the complete feature volume. The effectiveness of the proposed method is validated by comparing it with the state-of-the-art method quantitatively and qualitatively. Further ablation studies also show the need to adopt a hierarchical architecture to capture the multimodal distribution of possible shapes. 

\end{abstract}


\section{Introduction}
\label{sec:intro}

Perceiving 3D shapes is a challenging problem that has wide applications in robotics, AR/VR, and digital creations. 
However, when perceiving 3D shapes in real life, occlusion, truncation, and self-occlusion are always inevitable, leading to ambiguities on the 3D shapes. 
Recovering complete shapes from a partial observation is not a one-to-one mapping but a one-to-many mapping where the deterministic methods are neither suitable nor capable of providing multiple feasible solutions.
To address this ill-pose problem, a sound method needs to be able to model the multimodal distribution of possible shapes with diversities in both global structure and local details.

Recent research\cite{ONet, zhiqin2019imnet, park2019deepsdf} on 3D shape reconstruction has achieved significant progress by learning a continuous function of the 3D surface. This representation of the function space avoids direct discretization of the 3D space and the massive use of 3D convolution, leading to more efficient learning and a much smaller memory footprint. More recently, implicit functions with local priors~\cite{chibane20ifnet, ConvONet} have demonstrated their ability to capture fine-grained details and achieve better generalization across instances and classes. Moreover, this knowledge can be shared across categories and instances, since local latent codes just model local surface neighborhoods that could be shared across different instances. However, the local shape representation might be highly redundant as each local latent code prior could only capture a local region (typically a small voxel).

\begin{figure}[t!]
    \centering
    \includegraphics[width=0.8\linewidth]{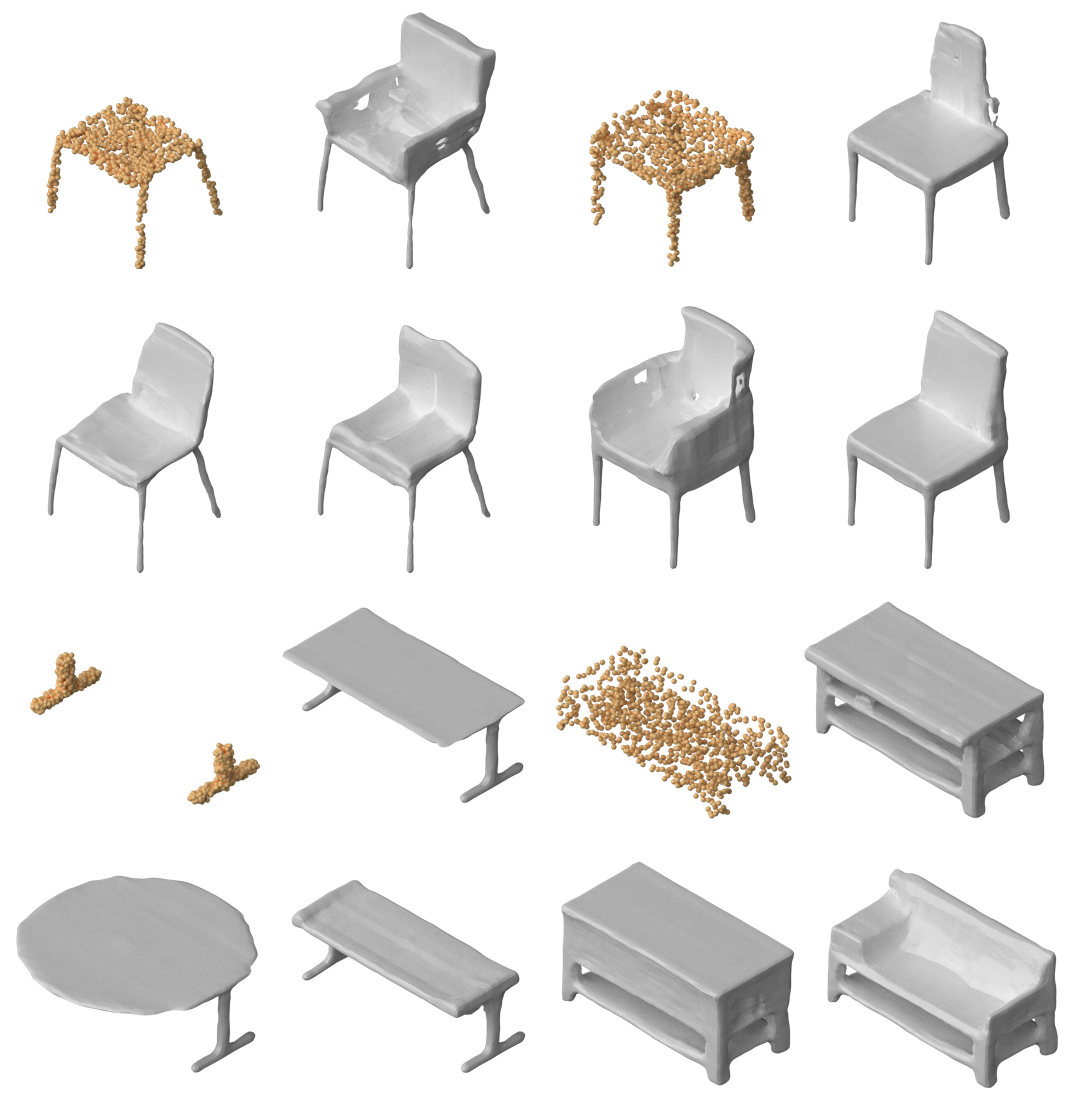}
    \caption{Given a partial point cloud as input (rendered in yellow color), our method could efficiently generate realistic and pluralistic completion results with a single feedforward step.
    }
    \label{fig:teaser}
\end{figure} 

Regression-based approaches, usually with a global latent code, tend to generate blurred results because they do not capture the multimodal distribution of possible shapes. 
The deterministic method~\cite{ONet, ConvONet, zhiqin2019imnet} with local implicit functions could perform well when the input point cloud is sparse but evenly sampled on the surface. 
However, these methods do not generate realistic or pluralistic reconstructions when the input shape is incomplete or ambiguous because their architectures do not explicitly model the multimodal distribution of the output. Some methods~\cite{park2019deepsdf, liu2022shadows} address this issue by optimizing the latent space of 3D shapes and trying to find latent code that could possibly match the input partial observation.
However, the optimization on the latent space mostly operates on the global latent code because it is hard to use partial observations to provide supervisions on the regions that do not contain input points. Therefore, a generative model that could potentially infer feasible shapes from partial observation is necessary to model this one-to-many mapping problem with high-quality details.

Several methods~\cite{yan2022shapeformer, zhang20223dilg, zhang2021gca} formulate this problem as an autoregressive generation of local latent codes conditioned on the latent code of known regions. Such methods need to apply the model hundreds of times in order to obtain a shape reconstruction of a single shape. Other methods \cite{wu2020multimodal} that use a global latent code could generate complete shapes in a single pass, but they typically cannot achieve a performance similar to that of an autoregressive method due to the limitations of the global latent code and less randomness introduced during the inference of possible shapes.

We propose a hierarchical variational autoencoder (VAE) architecture with a representation of canonical factors over the feature volume to address the problem of probabilistic shape completion from partial observations.  
Unlike previous methods that attempted to tackle this issue with an autoregressive model, our model starts from a global latent variable that encodes the global semantics of the input shape and estimates local latent variables for the feature field in a coarse-to-fine manner. 

A recent study~\cite{child2020very} on generative models indicates that hierarchical VAE architectures should be able to achieve performance similar to that of the autoregressive method with the same model complexity because autoregressive methods can be considered as a special hierarchical VAE where each stochastic layer takes the whole sequence of latent codes as input and only estimates a single latent vector, that will be appended to the input to the last layer.
Compared to autoregressive methods, general VAE architectures could be more flexible and expressive since the approximate posterior could perform probabilistic inference on the entire sequence of latent variables instead of only estimating the last latent variable\cite{child2020very}.
This suggests that hierarchical VAE could outperform the autoregressive method on image generation if the VAE model is deep enough, meaning that VAE models with multiple stochastic layers could potentially model the complex shape generation problem in a single feedforward network. 

We formulate the problem of shape completion as the completion on the feature volume of a local implicit function. By completing shapes in latent space, we can exploit the expressiveness of local implicit function. However, directly estimating local latent codes is highly inefficient because the number of local latent code grows cubically with the "resolution" of the feature volume. We notice the redundancy in the representation and propose to use a more compact representation, canonical factors, as the output of our probabilistic model.
Such factorization has already demonstrated its success in novel view synthesis, but was only used as a part of the optimization variables. We propose to use a network to estimate this local latent code. With the locality nature of the canonical rank representation, we could use multiple tiny multilayer perceptrons (MLPs) to gradually increase the level of details while avoiding the use of massive fully connected layers in the estimation. This further reduces the computational cost of our method. 

Both a quantitative and a qualitative study show that a single feedforward network is also capable of efficiently modeling this complex conditional distribution and generating 3D shapes with high fidelity and diversity. We conduct further ablation studies to show that our hierarchical design is critical to obtain reconstructions with high fidelity and diversity.

Our contributions can be summarized as follows:
\begin{compactitem}

    \item We propose a novel method for 3D shape completion that could achieve comparable or even better results than the autoregressive method while using much less computational resources.
    \item We evaluate our model with a common 3D shape dataset and benchmarks on shape completion and reconstruction. Our model demonstrates superior results both in terms of diversity and fidelity.
    \item We perform ablation studies showing that hierarchical design is critical in producing diverse shape completion results. 
\end{compactitem}

\section{Related Work}

\paragraph{Shape Completion}
Classical approaches focus on reconstructing surfaces from local cues by smoothing the point cloud input.
Hope~\etal~\cite{hoppe1992surface} reconstruct the surface by estimating the normal of the input points and creating the surface by synthesizing the tangent plane of the points. Horung~\etal~\cite{hornung2006robust} estimate the surface around the point cloud input using graph-cut-based energy minimization. 
More classical work can be found in this survey\cite{berger2017survey}. However, most of these methods do not aim to reconstruct surfaces that are not associated with point cloud input, as they cannot leverage categorical and semantic priors, except for some data-driven approaches that retrieve or composite from existing shapes~\cite{pauly2005example, shen2012structure}. 
Recent advances in deep learning and 3D shape representation learning have boosted research in this field significantly. Various methods have been used to generate point cloud output with generative adversarial networks (GAN)~\cite{wu2020multimodal, chen2019unpaired, gan_inv} normalizing flow~\cite{yang2019pointflow}, diffusion models~\cite{pvd}, other generative models\cite{achlioptas2018learning, sun2020pointgrow}, and  deterministic methods~\cite{topnet2019,yuan2018pcn, yu2021pointr}. However, these methods could only predict point clouds with a fixed number of points and do not provide surface and topology.  3D-EPN~\cite{dai2017complete} complete 3D shape with 3D CNN encoder and object retrieval. AutoSDF~\cite{autosdf2022} uses a Vector-Quantized VAE (VQ-VAE) as an encoder and decoder for 3D shapes and predicts the features of the invisible part autoregressively with a transformer~\cite{vaswani2017attention}. ShapeFormer~\cite{yan2022shapeformer} also used VQ-VAE and a transformer to complete the shape, but predicts a local implicit field instead of directly estimating the TSDF values.  Cheng~\etal\cite{cheng2022autoregressive}  uses autoregressive modeling with VQ-VAE but reconstructs point clouds instead. 3DILG~\cite{zhang20223dilg} extends the VQ-VAE and transformer autoregressive estimation pipeline with irregular feature grid representations. Zhang~\etal\cite{zhang2021gca} uses Generative Cellular Automata, another autoregressive method that gradually estimates the occupancy of the voxel around existing voxels, to complete pluralistic 3D shapes.

\paragraph{Variational Autoencoders}
VAE~\cite{kingma2013auto} is a generative approach that models the joint distribution $p(x, z)$ which is factorized as $p(x, z) = p(z)p(x|z)$  where $p(z)$ is the prior distribution over the latent space and $p(x|z)$ is a stochastic decoder that decodes the latent code $z$ into the data sample ${\hat x}$. Since the posterior distribution $p(z|x)$ is typically intractable, an approximate posterior distribution $q(z|x)$ is learned using the reparameterization trick~\cite{kingma2014semi, rezende2014stochastic}. Both the encoder and decoder can be modeled by a neural network. 
VAE can be extended with using hierarchical approximate posterior encoder and decoder~\cite{sohn2015learning,kingma2016improved,vahdat2020nvae, child2020very} to improve the fidelity of the generation. VAE can also model the conditional generative task~\cite{sohn2015learning, zheng2021pluralistic, wan2021high, ivanov2018variational, harvey2022conditional} by modeling  $p(x|y)$.

\paragraph{Tensor Decomposition.}
Tensor decomposition has been a long-standing problem that has applications in various fields~\cite{kolda2009tensor}. Tensor rank decomposition or canonical polyadic decomposition~\cite{carroll1970analysis, harshman1970foundations} is an efficient way to represent tensors by generalizing singular value decomposition (SVD) to tensors. Tensor rank decomposition has been applied to some vision tasks, such as representing 3D neural fields~\cite{Chen2022ECCV}. 
The Tucker decomposition is another generalization of the matrix SVD but the number of parameters is exponential to the dimension $d$. Block term decomposition is a recent combination of tensor rank decomposition and Tucker decomposition~\cite{block_term}, and has been applied to many vision and learning tasks~\cite{ye2020block, ben2019block, ye2018learning}.  Oseledets~\etal~\cite{tensor_train} proposed a tensor train decomposition based on low-rank approximation of auxiliary unfolding matrices, which has been applied to  vision and learning tasks\cite{boyko2020tt, usvyatsov2021cherry}. We focus on the tensor rank decomposition because canonical factors can be efficiently estimated through multiple tiny MLPs.

\section{Technical Approach}

\begin{figure*}[t]
	\centering
	\includegraphics[width=2.0\columnwidth,]{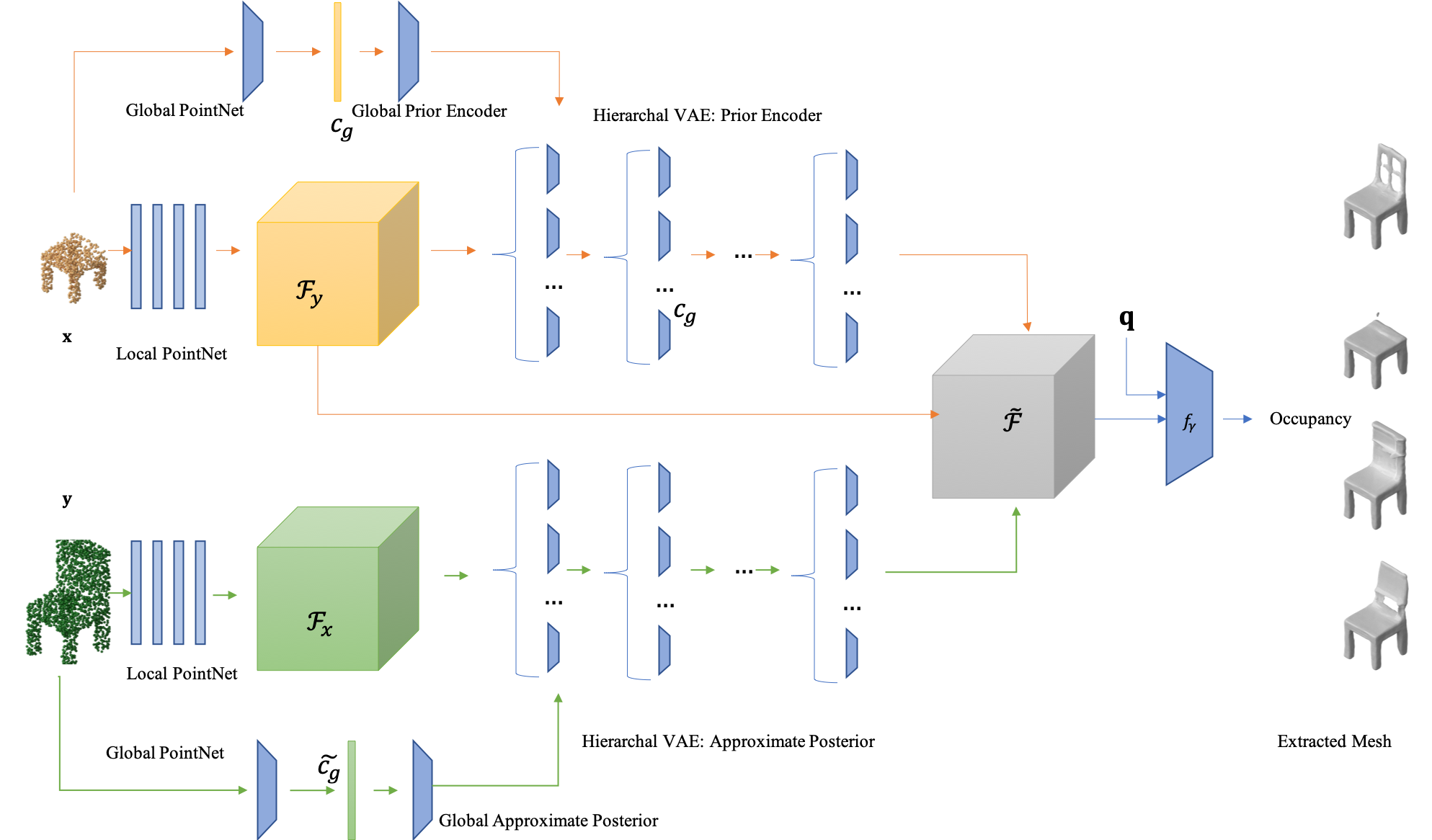}
	\vspace{-2mm}
	\caption{{\bf Overview of the proposed approach.}
	We adopt a hierarchical VAE architecture to complete the features on the missing parts. The branch of the approximate posterior is marked in green  and only used for training, whereas the branch of the prior encoder is marked in yellow and will be used during testing. Given a pair of partial and complete 3D shapes, our model first encode the point cloud into a global feature vector and local latent fields and then uses a hierarchical VAE to complete the local feature volume. With the feature volume $\Bar{\mathcal{F}}$ of the complete shape, realistic shapes can be decoded by querying the local implicit functions with marching cube algorithm.
}
\label{fig:pipeline}
\vspace{-5mm}
\end{figure*}

Given a partial point cloud observation $\mathcal{P}\in\mathbb{R}^{N\times 3}$, our aim is to reconstruct multiple plausible 3D meshes that match the observation. We first encode the input partial observation into a local feature grid and then estimate the factors of tensor rank decomposition that can be used to reconstruct the 3D feature volume of the complete shape. 

The complete shape can be decoded from the reconstructed 3D feature volume with a local implicit function decoder. 
We start by providing information on the local implicit field encoder and decoder in Sec.~\ref{sec:impl_func} and then introduce the tensor rank decomposition that enables efficient reconstruction for the feature field of the reconstructed shape (Section~\ref{sec:tensor_repr}).
In the following, we present our hierarchical VAE architecture in Sec.~\ref{sec:hvae}, and details about the training and testing of the proposed model (Section~\ref{sec:impl_details}). The overview of our pipeline can be found in Fig.\ref{fig:pipeline}.

\subsection{Point cloud Encoder and Implicit Function Decoder} \label{sec:impl_func}

To address the ill-posed challenge of multimodal shape completion, we need to model both global semantic information and the local structure of the input point cloud $\mathcal{P}$. 
Following the Occupancy Network~\cite{ONet} and the Convolutional Occupancy Network~\cite{ConvONet}, we use PointNet~\cite{PointNetPP} as the encoder of our model.
A global feature vector $c \in \mathbb{R}^n$ is encoded by a PointNet with global pooling. 
A shallow PointNet with local pooling is used to obtain per-point features. 
A feature volume $\mathcal{F}^{H\times W \times D \times d}$, where $d$ is the feature dimension, is reconstructed by pooling features from the points that fall into its voxels.
Thus, the input point cloud $\mathcal{P}$ can be mapped to features by global and local PointNet encoders:
\begin{align}
\label{eq:global_pp}    f_g : \mathcal{P} \rightarrow c\\
\label{eq:local_pp}    f_l: \mathcal{P} \rightarrow \mathcal{F}
\end{align}

The global feature $c$ and the local feature volume $\mathcal{F}$ will be used as contextual information for our generative model, which will reconstruct a complete feature volume $\Tilde{\mathcal{F}}$. 
Given such a feature volume for the complete shape, the complete mesh can be extracted by predicting the occupancy probabilities on the field with an implicit function and applying a marching cube algorithm~\cite{michalkiewicz2019miset}. For each query point ${\bf q} \in \mathbb{R}^3$, the feature vector $\varphi(\bf{q}, \Bar{\mathcal{F}})$ associated with the input point can be sampled from the feature field $\Bar{\mathcal{F}}$. 
The probabilities of the occupancy of the point can then be predicted by a fully connected network\cite{ConvONet}:
\begin{equation}\label{eq:impl_func}
    f_\gamma( {\bf q}, \varphi({\bf q}, \Bar{\mathcal{F}}) ) \rightarrow  [0, 1]
\end{equation}

We refer the reader to the original paper\cite{ConvONet} for details of the implicit function network.
With PointNet encoders and the implicit function decoder, the problem of shape completion can be reduced to estimating the complete feature field $\Bar{\mathcal{F}}$ from the feature field $\mathcal{F}$ of the partial point cloud. 

\subsection{Tensor Rank Decomposition} \label{sec:tensor_repr}
A key observation for the local feature field is that the local features should be very similar for regions with a similar surface, even if they are from different objects. We could also observe that the local isosurface for most real-life objects is simple and repetitive. 
From these observations, the estimation of the local feature volume $\Bar{\mathcal{F}} \in \mathbb{R}^{H\times W\times D\times d}$ can be considered as the estimation of a low-rank tensor that can be factorized into more compact representations. 
Thus, we use the tensor rank decomposition~\cite{carroll1970analysis, harshman1970foundations} to decompose the feature volume. The element of a $d$-dimensional tensor $\mathbf{A}$ can be represented as~\cite{tensor_train}: 

\begin{equation}
    \mathbf{A}(i_1, i_2, ..., i_d) = \sum_{r=1}^R U_1(i_1, \alpha) U_2(i_2, \alpha)...U_d(i_d, \alpha),
\end{equation}
where $R$ is the \textit{tensor rank} and matrices $U_k=[U_k(i_k, \alpha)]$ are \textit{canonical factors}.

As we aim to reconstruct feature volume of 3D space, we need to perform tensor rank decomposition on a 4D tensor with 3 spatial dimensions and a feature channel $\Bar{\mathcal{F}} \in \mathbb{R}^{H\times W\times D\times d}$. The extended tensor rank decomposition for this 4D tensor can be written as:
\begin{equation}
    \Bar{\mathcal{F}} = \sum_{r=1}^R v_r^1 \circ v_r^2 \circ v_r^3,
\end{equation}
where $\circ$ denotes the outer product and $v_r^1\in \mathbb{R}^{H\times d},  v_r^2 \in \mathbb{R}^{W\times d}, v_r^3 \in \mathbb{R}^{C\times d}$ are the canonical factors of the $r$-th tensor component and $R$ is the tensor rank. Determining the tensor ranks, the number of groups of canonical factors, is an NP-complete problem~\cite{haastad1989tensor}. Therefore, we use a fixed canonical rank in our experiments.  Each element in the tensor volume can be calculated as the sum of point-wise multiplication among the corresponding locations
\begin{equation}
    \Bar{\mathcal{F}}_{i j k t} = \sum_{r=1}^R v_{rt}^{1, i} \cdot v_{rt}^{2, j} \cdot v_{rt}^{3, k}
\end{equation}
where $v_{rt}^i$ denotes the $t$-th feature value of the feature vector of the $r$ -th tensor component with index $i$ on the first dimension. This can also be considered as a concatenation of $d$ tensors constructed by the original 3D tensor factorization~\cite{Chen2022ECCV}.

With tensor rank decomposition, we can estimate the complete shape of the object by estimating the factors of the complete feature volume. Therefore, the feature field for the complete shape can be obtained by combining the input feature field $\mathcal{F}$ and the predicted local features
\begin{align}
    \Bar{\mathcal{F}} =\big( \sum_{r=1}^R v_r^1 \circ v_r^2 \circ v_r^3\big) \oplus\mathcal{F}
\end{align}
where $\oplus$ is the concatenation of features to preserve the fidelity for the visible part of the partial object. 
Thanks to this efficient representation, we can reduce the number of local latent vectors for the feature field from $\mathcal{O}(n^3)$ to $\mathcal{O}(n)$.

\subsection{Hierarchical VAE for Factor Estimation} \label{sec:hvae}
Given the encoded partial observation $\bf{x}=\{c_g, \mathcal{F}\}$, we need to model the conditional probability distribution $p(\bf{y}|\bf{x})$ where $\bf{y}$ is the feature volume that $\Bar{\mathcal{F}}$ can be decoded into a complete mesh $\mathcal{M}$.
We could formulate this problem as a CVAE~\cite{sohn2015learning}. That is, the approximate posterior $q_\phi (\bf{z}|\bf{x}, \bf{y})$ or the prior $p_\theta(\bf{z}| \bf{x})$ encodes the input features in a latent variable $\bf{z}$, which will then be decoded to the complete feature volume $\bf{y}$ by a generator $p_\theta(\bf{y}| \bf{x}, \bf{z})$ at training or testing time, respectively.
The {\it evidence lower bound} (ELBO) of the CVAE model~\cite{sohn2015learning} can be written as: 
\begin{equation}\label{eq:elbo}
     \begin{aligned}
        \log p_\theta(\bf{y}|\bf{x}) \geq & -D_\text{KL}\big( q_\phi({\bf{z}|\bf{x}, \bf{y}}) || p_\theta({\bf z|\bf{x}})\big)\\
& + \mathbb{E}_{q_\phi(\bf{z}| \bf{x}, \bf{y})}\big[\log p_\theta({\bf{y}|\bf{x}, \bf{z}}) \big]
    \end{aligned}   
\end{equation}
where $D_\text{KL}(.||.)$ is the Kullback-Leibler divergence. The network parameters $\phi$ and $\theta$ can be jointly optimized by maximizing this ELBO.

A straightforward way to complete the partial shape is to sample a global latent variable $\bf{z}$ that can be used by an implicit function decoder, but we argue that a global latent vector is not efficient in capturing the multimodal nature of the details for the unseen part. 
Instead, we use a set of latent vectors $\{ z^{(0)}, z^{(1)}, ... , z^{(n)} \}$ so that each vector is responsible for a local region of the feature volume as the latent variable $\bf{z}$. 
However, global contextual information is still helpful to generate realistic complete shapes. Therefore, we adopt a hierarchical VAE~\cite{sonderby2016ladder} structure with conditional input to achieve expressive and pluralistic shape completions. That is, the latent $\bf{z}$ is modulated as a set of latent code ${\bf{z}}_i,\  i=1, ... , L$, where $L$ is the number of layers. The latent variable ${\bf z}_i$ depends on the latent code in the lower stochastic layers with a lower resolution but a larger receptive field:
\begin{align}
    & p_\theta({\bf{z | x}}) = p_\theta({\bf z}_1 | {\bf c}_x)\prod_{i=1}^{L-1} p_\theta({\bf{z}}_{i+1} | {\bf z}_{i}, {\bf x}_i)\\
    & q_\phi({\bf{z | x, y}}) = p_\theta({\bf z}_1 | {\bf c}_x, {\bf c}_y)\prod_{i=1}^{L-1} q_\phi({\bf{z}}_{i+1} | {\bf z}_{i}, {\bf x}_i, {\bf y}_i)
\end{align}
where ${\bf c}_x$ is the global conditional vector of the partial shape of Eq.~\ref{eq:global_pp}, and ${\bf x}_i, {\bf y}_i$ is obtained by average pooling the features with the designated resolution of the layer $i$ from the feature volume ${\mathcal{F}_x}$ and $\Bar{{F}_y}$ from Eq.~\ref{eq:local_pp}.
This distribution could be modeled by an autoregressive model, but we suggest that a well-designed feedforward network with multiple stochastic layers could efficiently model this distribution.
Latent variables $ {\bf z}_L = \{ z^{(0)}_L, z^{(1)}_L, ... , z^{(n)}_L \}$ from the last stochastic layer will be used to obtain the tensor rank decomposition factors of the feature volume for the complete shape. 
This design can also be interpreted as a coarse to fine architecture for the shape completion, where the model first estimates a rough distribution of the complete shape, and more details would be introduced in the later layers. 
Hence, the shape completions could be pluralistic with different local structures but are still coordinated by the global context. 

\begin{figure}[h]
    \centering
    \includegraphics[width=1.0\linewidth]{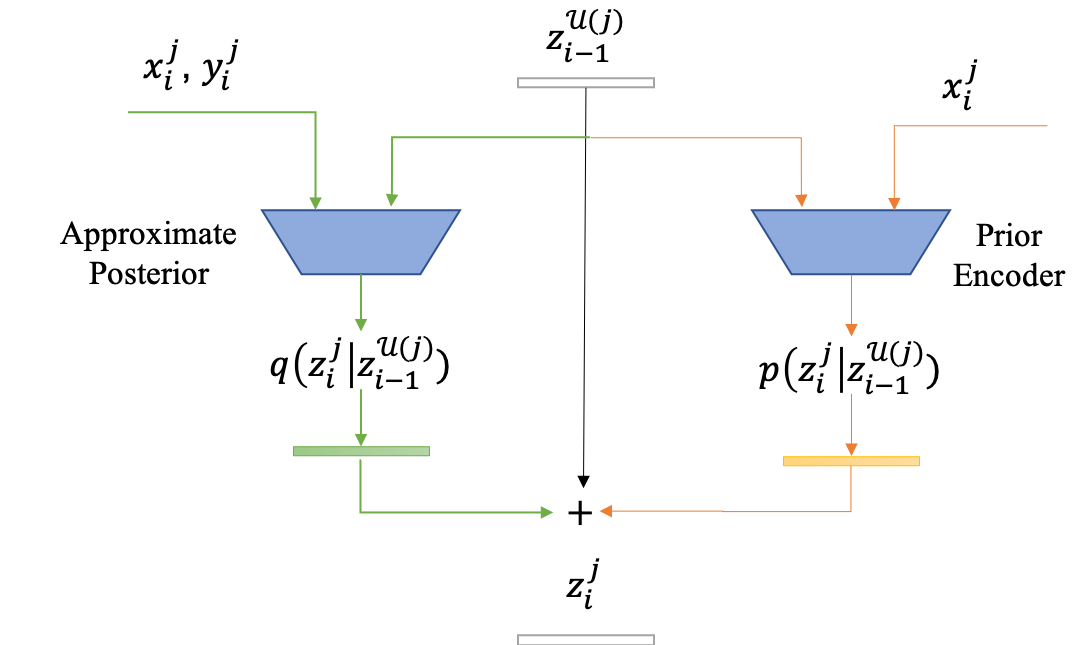}
    \caption{\textbf{Diagram of Our Stochastic Layer}. The green branch is used at the training time and the KL divergence loss between $q(z_i^j|z_{i-1}^{\mathcal{U}(j)})$ and $p(z_i^j|z_{i-1}^{\mathcal{U}(j)})$ is optimized. At the test time, the orange branch will be used to apply residual on the latent code from the last layer. 
    }
    \label{fig:layer}
\end{figure} 

As we seek to estimate local latent code with fine-grained details, we could further decouple the conditional distribution $p({\bf z}_{i+1}| {\bf z}_i)$ so that each latent variable is only conditioned on the latent variables from the last stochastic layer representing the same region. 

\begin{align}
    & p({\bf z}_{i+1}| {\bf z}_i)= p(\{z_{i+1}^{(1)}, ... , z_{i+1}^{(n)} \} \big| \{z_i^{(1)}, ... , z_i^{(n)} \}) \\
    & p_\theta({\bf z}_{i+1}^j | {\bf z}_i^{\mathcal{U}(j)}) = \mathcal{N}(z_{i+1}^j|\mu_{i,j}(z_{i}^{\mathcal{U}(j)}),  \sigma_{i,j}^2(z_{i}^{\mathcal{U}(j)}))   
\end{align}


where $\mathcal{U}(j)$ is the index of the neighboring latent variable in the last resolution, and $\mu_{i, j}$ and $\sigma_{i, j}$ are functions of the parameters of the Gaussian distribution $\mathcal{N}$. Conditional variables $x_i$ and $y_i$ are ignored in this equation. This enables us to use multiple tiny MLPs to efficiently train and infer with our model. A diagram of our stochastic layer in Figure~\ref{fig:layer} to better understand the behavior of our model during training at the test time.

\subsection{Training and Inference} \label{sec:impl_details}

Our implementation is done using PyTorch and the publicly available libraries. Detailed hyper-parameters and network architecture can be found in the supplementary material.

All the parameters of our model are trained end-to-end. The training objective can be written as:
\begin{align} \label{eq:loss}
    & \mathcal{L} = \mathcal{L}_\text{recon} + \lambda \mathcal{L}_\text{KL}\\
    & \mathcal{L}_\text{recon} = \sum_{{\bf q} \in Q}\text{BCE}( f_\gamma( \bf{q}, \varphi(\bf{q}, \Bar{\mathcal{F}}) ), \mathbf{o}(q))\\
    & \mathcal{L}_\text{KL} = D_\text{KL}\big( q_\phi({\bf{z}|\bf{x}, \bf{y}}) || p_\theta({\bf z|\bf{x}})\big)
\end{align}
where $\text{BCE}(., .)$ is the binary cross entropy loss, and $Q \in \mathbb{R}^{N\times 3}$ are the query points and ${\bf o}(.)$ is the ground truth occupancy function, and $\lambda$ is an annealing weight that gradually increases as training progresses.

At training time, latent variables $\bf{z}$ are sampled from the approximate posterior distribution ${\bf z} \sim  q_\phi({\bf{z}|\bf{x}, \bf{y}})$. Whereas at inference time, the approximate posterior $q_\phi$ is discarded and the latent code $\bf{z}$ in each stochastic layer will be sampled from the prior distribution ${\bf z} \sim {p}_\theta({\bf z|\bf{x}})$.

\section{Experiments}
In this Section, we present the empirical evaluation of our approach. First, we describe the datasets used for training and evaluation (Sec. \ref{sec:dataset_eval}). Then we focus on
the quantitative evaluation and ablation study of our approach (Sec.\ref{sec:comp_sota} and \ref{sec:ablation}), and
finally, we present more qualitative results (Sec.\ref{sec:qualitative}).

\subsection{Datasets and Evaluation Protocols}\label{sec:dataset_eval}

\paragraph{Train and Test Data}

To demonstrate the effectiveness of our method,  we trained our models on objects from the ShapeNet dataset~\cite{shapenet2015} to perform quantitative and qualitative evaluations of object completion and surface reconstruction. For shape completion on a single object, we use the \textit{train/val/test} split from 3D-EPN following previous work~\cite{dai2017complete, wu2020multimodal, autosdf2022}.

Following the evaluation protocol of AutoSDF~\cite{autosdf2022}, we created two types of partial point cloud input: 

\begin{itemize}

    \item \textit{Bottom}: points on the bottom half of the object are sampled as partial observation
    \item \textit{Octant}: points are sampled from the left and bottom half of the complete shape as partial observation.
\end{itemize}

For all experiments, we trained a single model on all 13 categories for each type of partial point cloud input. 
The categories of chair, airplane, and table are used as testing categories. 
1024 points are sampled for the partial point cloud, and 2048 points are sampled as the complete point cloud. 
Query points are sampled uniformly within the unit cubic to train the implicit function.

For auto-encoding experiments, we use the \textit{train/val/test} split from 3D-R2N2\cite{choy20163d} and the training data pre-processed by Mescheder~\etal\cite{ONet}. Following the training protocol of ConvONet\cite{ConvONet}, 3000 points are sampled on the surface of the object as the input point cloud. We trained and evaluated in all 13 categories. 2048 points are sampled as the query for the training of the implicit function. 

 \paragraph{Evaluation Metric}
 We use the evaluation metrics proposed by MPC~\cite{wu2020multimodal} for the evaluation on shape completion. The evaluation metrics take into account both fidelity and diversity. 
 \textit{Unidirectional Hausdorff Distance} (UHD) is used to evaluate the fidelity of the shape completion, which is described as the average distance between the input partial observation and the model output. 
 The diversity metric is measured by \textit{total mutual difference} (TMD), which is the average pairwise distances between all completion results for each partial observation. Ten samples are generated for each partial observation to calculate the UHD and TMD.

Following\cite{ONet, ConvONet} , we use \textit{intersection over union} (IoU), $L_1$ Chamfer Distance (Chamfer-$L_1$), Normal Consistency(Norml C.), and F-Score as evaluation metrics for auto-encoding experiments. We refer the reader to  Peng~\etal\cite{ConvONet} for the details of the evaluation metrics.

\subsection{Comparison with the state-of-the-art}\label{sec:comp_sota}
\begin{table}\centering
	\begin{tabular}{c c c c c}
		\toprule
		\multirow{2}{*}{Method}  & \multicolumn{2}{c}{Bottom}  &  \multicolumn{2}{c}{Octant} \\

        & UHD $\downarrow$ & TMD $\uparrow$ & UHD $\downarrow$ & TMD $\uparrow$\\
        \midrule
        MPC\cite{wu2020multimodal} & 0.0627 & 0.0303 & 0.0579 & 0.0376\\
        PointTr \cite{yu2021pointr} & 0.0572 & N/A & 0.0536 & N/A \\
        AutoSDF\cite{autosdf2022} & 0.0567 & 0.0341 & 0.0599 & 0.0693\\
        \midrule
        Ours & 0.0400  & 0.0413& 0.0397 & 0.0669 \\
		\bottomrule
	\end{tabular}
	\caption{\textbf{Quantitative comparison on Shape Completion.} Numbers for other methods are obtained from Mittal~\etal\cite{autosdf2022} PointTr\cite{yu2021pointr} is not evaluated for diversity metric because it does not generate multiple reconstructions from the input.}
	\label{tb:shapenet_comp}
\end{table}

\begin{table}\centering
	\begin{tabular}{c c c}
		\toprule

        & \# params  & \# forward-pass  \\
        \midrule
        ShapeFormer\cite{yan2022shapeformer} & 323M  &  217 \\
        AutoSDF\cite{autosdf2022} & 67.3M &  256 \\
        \midrule
        Ours & 15.9M &  1 \\
		\bottomrule
	\end{tabular}
	\caption{\textbf{Comparison on the Computational Cost} This does not include parameters that are not retained during testing (\eg approximate posterior network for VAE architectures.). \# inference indicates the number of forward-pass needed to generated a single sample. Shape-former uses a variable length of sequence, 217 is the average sequence length.}
	\label{tb:compute_cost}
\end{table}

We compare our method with recent state-of-the-art methods~\cite{wu2020multimodal, yu2021pointr, autosdf2022} on 3D object shape completion.
As can be seen in Table~\ref{tb:shapenet_comp} our model outperforms the previous methods in both fidelity and diversity metrics when the bottom half of the point cloud is the input. Our model also achieves comparable results when input observation is extremely limited.  
In addition, the analysis on computational cost can be found in Table.~\ref{tb:compute_cost}.
Thanks to the efficient representation powered by the tensor rank decomposition and the hierarchical VAE architecture, our model achieves state-of-the-art performance with much less computational cost in terms of the number of parameters.

\begin{figure*}[t]
    \centering
    \includegraphics[width=0.8\linewidth]{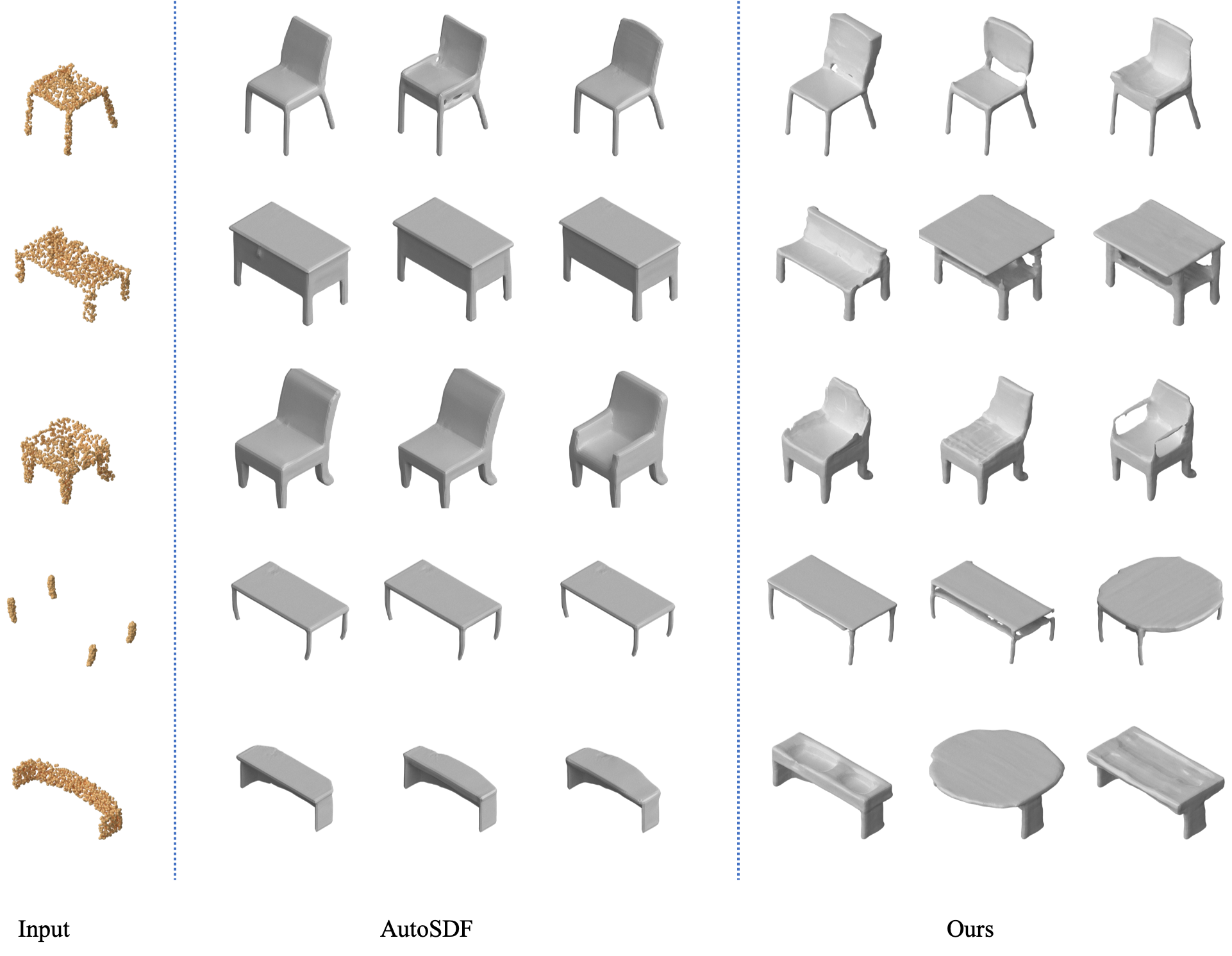}
    \caption{\textbf{Comparison with state-of-the-art}. We compare our shape completion result with the state-of-the-art on the test set of the ShapeNet dataset. As can be seen, our model completes diverse shape on the details and faithfully preserve the details for the input partial shape. The results from AutoSDF~\cite{autosdf2022} are generated from the checkpoint provided by the authors.}
    \label{fig:compare}
\end{figure*}

We also qualitatively compare our method with the latest state-of-the-art~\cite{autosdf2022} in Figure~\ref{fig:compare}.

\subsection{Ablation Study} \label{sec:ablation}
\begin{table}\centering 
	\begin{tabular}{c c c c }
	
		\toprule
		  & ONet\cite{ONet} &  ConvONet\cite{ConvONet} & Ours \\
        \midrule
        IoU $\uparrow$ & 0.700 & 0.851 & 0.848\\
        Chamfer-$L_1$ $\downarrow$  & 0.011 & 0.005  & 0.005\\
        Normal C. $\uparrow$  & 0.877 &  0.926 &  0.925\\
        F-Score $\uparrow$ & 0.675  &  0.917    & 0.912\\
		\bottomrule	

	\end{tabular}
	\caption{\textbf{Auto-encoding results for objects in ShapeNet.} All the models are trained with 200k iterations. We use the hyperparameters from the original paper\cite{ONet} to train ONet and use the same feature dimensions and feature volume resolution to train ConvONet and our method. }
	\label{tb:shapenet_ae}
\end{table}
As our method is built on the local point cloud encoder and the implicit function decoder of ConvONet~\cite{ConvONet}, we wish to validate that our tensor rank decomposition does not obstruct the performance of the original ConvONet. 
Therefore, we perform evaluations on ShapeNet~\cite{shapenet2015} with the same settings as ConvONet.
We ignore the prior encoder of our hierarchical VAE because the input just contains complete point cloud in this metric and use the approximate posterior to encode the latent vector directly instead of sampling from the Gaussian distribution.  

As can be seen in Table~\ref{tb:shapenet_ae}, our model achieves a similar performance to ConvONet, indicating that our representation does not compromise the quality of the reconstruction.

\begin{table}\centering
	\begin{tabular}{c c c}
		\toprule

        & TMD $\uparrow$ & UHD $\downarrow$ \\
        \midrule
        Global & 0.0168 & 0.0423 \\
        Global - Factors & 0.0103 & 0.1580 \\
        Local  &  0.0168  & 0.0423  \\
        Hierarchical & \textbf{0.0413} & \textbf{0.0400} \\
		\bottomrule
	\end{tabular}
	\caption{\textbf{Ablation Study on Shape Completion.} We conduct ablative studies on the ShapeNet dataset\cite{shapenet2015} with both fidelity (UHD) and diversity (TMD) metrics. The point cloud from the bottom of testing objects are sampled as input. All the methods with local implicit field are trained with the same resolution. }
	\label{tb:shapenet_abl}
\end{table}
To study the effectiveness of our hierarchical VAE architecture, we compare our hierarchical VAE with other VAE implementations while keeping other parts of the model the same:
\begin{itemize}

    \item \textit{Global}: the model that uses only global latent code with the same configuration as the occupancy network encoder and decoder. 
    \item \textit{Global - Factors}: this model also uses a global VAE to predict a single latent vector. However, this vector is used to first decode the canonical factors for the feature volume and then a local implicit function~\cite{ConvONet} is used to predict the occupancy value with the reconstructed feature volume. 
    \item \textit{Local}: this model only uses a single stochastic layer with the highest resolution to predict the canonical factors of the complete feature field. No global latent codes are used as conditional variables for this local VAE model. 
\end{itemize}

The quantitative comparison in Table.~\ref{tb:shapenet_abl} shows that the hierarchical VAE architecture is critical to generate pluralistic shape completion. Completion from both global and local variants could maintain high fidelity, whereas only hierarchical VAE could achieve superior results on the diversity metric. 
One possible interpretation is that the hierarchical VAE architecture could perform conditional sampling multiple times when generating more detailed latent vectors, whereas VAE with a single stochastic layer, whatever is used to encode a global latent code or a set of local latent codes, could only introduce random sampling once. This could also explain why autoregressive methods could achieve promising results in multimodal shape completion.

\subsection{Qualitative Results}\label{sec:qualitative}

\begin{figure*}[h]
    \centering
    \includegraphics[width=0.95\linewidth]{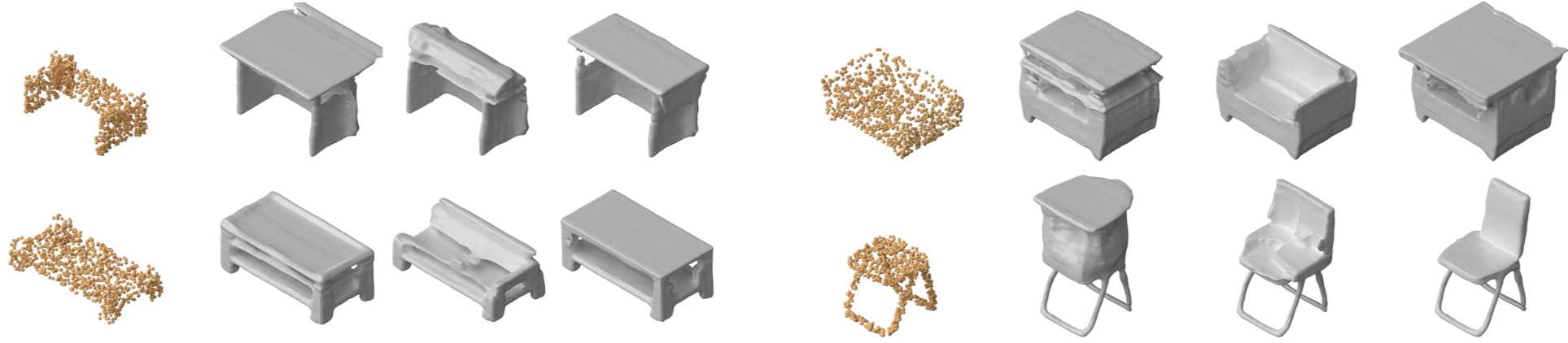}
    \caption{\textbf{Qualitative Study on Our Approach}. We visualize our shape completion results on the test set of ShapeNet~\cite{shapenet2015}. The input point cloud is rendered in yellow and is sampled from the bottom half of the object. }
    \label{fig:qual}
\end{figure*} 

As demonstrated in Figure~\ref{fig:qual}, our model could capture the multimodal distribution of the complete shapes while preserving the details on the visible part. It is surprising to find that our model could complete shapes with different categories when the partial point cloud input could not reveal the category of the object.
However, we could also observe artifacts in our shape completion, probably because the tiny MLPs are independent of each other and could have disagreements when estimating the local surface even though they are conditioned on the same latent variables in previous stochastic layers. This is a trade-off compared to the autoregressive methods in which each latent variable is estimated by conditioning all other latent variables.


\section{Conclusion and Limitations}
We studied the problem of probabilistic shape completion from partial observations.
By formulating the shape completion problem as a tensor completion problem on the latent feature volume, we could take advantage of the existing point cloud encoder and local surface decoder.

With the powerful and efficient probabilistic model and compact representations, our method achieves better or comparable results on multimodal shape completion benchmarks with much fewer parameters and computational cost. 

However, our completion results sometimes experience incoherence on some detailed surfaces, which could potentially be resolved by improving the design of tiny MLPs network architecture to eliminate the disagreement on stochastic layers. Another limitation is that the current model has only been experimented with single-object completion. More research could be conducted to extend this method to dynamic objects or large-scale scene completion in the future. 

\newpage
{\small
\bibliographystyle{ieee_fullname}
\bibliography{egbib}
}

\end{document}